\documentclass{article}

\usepackage{microtype}
\usepackage{graphicx}
\usepackage{subfigure}
\usepackage{booktabs} 

\usepackage{hyperref}

\usepackage{amsmath}
\usepackage{amssymb}
\usepackage{mathtools}
\usepackage{amsthm}

\usepackage{algorithm}
\usepackage{algorithmic}

\usepackage{booktabs} 

\usepackage[english]{babel}
\usepackage{moresize}
\usepackage{amsmath}

\usepackage{balance}
\usepackage{comment}
\usepackage{paralist}
\usepackage{bm}
\usepackage{pgfplots}
\usetikzlibrary{pgfplots.dateplot}

\usepackage{flushend}
\usepackage[english]{babel}
\usepackage[latin1]{inputenc}
\usepackage{mathrsfs}
\usepackage{graphicx}

\usepackage{amssymb}
\usepackage{amsfonts}
\usepackage{url}
\usepackage{longtable}
\usepackage{rotating}
\usepackage{multirow}
\usepackage{mathrsfs}
\usepackage{subfigure}
\usepackage{enumitem}
\usepackage{adjustbox}
\usepackage{hyperref}
\usepackage{bbding}
\usepackage{threeparttable}
\usepackage{balance}
\usepackage[accepted]{icml2024}

\usepackage{amsmath}
\usepackage{amssymb}
\usepackage{mathtools}
\usepackage{amsthm}

\usepackage[capitalize,noabbrev]{cleveref}

\theoremstyle{plain}

\theoremstyle{definition}

\theoremstyle{remark}

\newcommand{\eg}{\textit{e}.\textit{g}.} 

\usepackage[textsize=tiny]{todonotes}

\def\model{FlashST}

\begin{document}

\twocolumn[
\icmltitle{FlashST: A Simple and Universal Prompt-Tuning Framework for \\ Traffic Prediction}




\begin{icmlauthorlist}
\icmlauthor{Zhonghang Li}{scut,hku}
\icmlauthor{Lianghao Xia}{hku}
\icmlauthor{Yong Xu}{scut,pzl}
\icmlauthor{Chao Huang*}{hku}
\end{icmlauthorlist}

\icmlaffiliation{scut}{South China University of Technology}
\icmlaffiliation{hku}{University of Hong Kong}
\icmlaffiliation{pzl}{PAZHOU LAB}

\icmlcorrespondingauthor{Chao Huang}{chaohuang75@gmail.com}


\icmlkeywords{Machine Learning, ICML}

\vskip 0.3in
]



\printAffiliationsAndNotice 

\begin{abstract}
The objective of traffic prediction is to accurately forecast and analyze the dynamics of transportation patterns, considering both space and time. However, the presence of distribution shift poses a significant challenge in this field, as existing models struggle to generalize well when faced with test data that significantly differs from the training distribution. To tackle this issue, this paper introduces a simple and universal spatio-temporal prompt-tuning framework-\model, which adapts pre-trained models to the specific characteristics of diverse downstream datasets, improving generalization in diverse traffic prediction scenarios. Specifically, the \model\ framework employs a lightweight spatio-temporal prompt network for in-context learning, capturing spatio-temporal invariant knowledge and facilitating effective adaptation to diverse scenarios. Additionally, we incorporate a distribution mapping mechanism to align the data distributions of pre-training and downstream data, facilitating effective knowledge transfer in spatio-temporal forecasting. Empirical evaluations demonstrate the effectiveness of our \model\ across different spatio-temporal prediction tasks using diverse urban datasets. Code is available at \href{https://github.com/HKUDS/FlashST}{https://github.com/HKUDS/FlashST}.

\end{abstract}

\section{Introduction}
\label{sec:intro}


Accurate traffic forecasting is a crucial objective in the domain of urban computing. It aims to precisely predict and analyze the dynamic transportation patterns within cities. The key goals of traffic prediction include supporting urban planning, enabling real-time monitoring and management of traffic, and contributing to the development of smart city applications~\cite{Guo2019attention,han2021dynamic,li2024urbangpt}.. The primary challenge lies in effectively modeling the intricate spatial and temporal correlations among different time intervals and geographical locations.

Neural network techniques have become a prominent approach for spatio-temporal prediction tasks, thanks to their remarkable capabilities in feature representation. Initially, researchers leveraged Recurrent Neural Networks (RNNs) \cite{yu2017deep} to capture temporal dependencies, and Convolutional Neural Networks (CNNs) \cite{zhang2017deep,yao2018deep} to model spatial correlations. Subsequently, Graph Neural Networks (GNNs) have emerged as effective models for capturing complex spatial dependencies, utilizing message passing mechanisms to learn the relationships among spatial units such as regions and road segments. In constructing the graph adjacency matrices, researchers have explored various factors. Some have considered static geographical distance \cite{li2018diffusion,yu2017spatio}, while others have incorporated time-aware region correlations \cite{han2021dynamic,li2022spatial}. Additionally, there have been efforts to learn the region-wise relevance in a data-driven manner \cite{wu2019graph, wu2020graph}.


Although the aforementioned methods have shown effectiveness, most current spatio-temporal prediction models struggle to generalize effectively when confronted with distribution shifts across diverse downstream datasets and tasks. In their methods, the assumption of an inconsistent distribution between training and testing data becomes a hindrance to accurate predictions in real-life urban scenarios with uncontrolled shifts from unseen spatio-temporal data. As illustrated in Figure 1, directly applying parameters learned from training on dataset A to test on dataset B can lead to suboptimal performance due to significant variations in spatio-temporal characteristics across different data distributions. Therefore, there is a need to enhance the generalization ability of spatio-temporal forecasting models by efficiently adapting them to handle such distribution shifts.

While there are potential benefits to enabling model adaptation for spatio-temporal prediction methods, several key challenges remain to be addressed. \emph{Firstly}, \textbf{C1}: efficiently distilling specific and complex spatio-temporal contextual information from the downstream tasks is essential. However, equipping pre-trained models with the ability to understand and incorporate the spatial and temporal characteristics of new domain data that is only accessible during testing is a formidable challenge. \emph{Secondly}, \textbf{C2}: there is often a significant distribution gap between the training and test datasets, especially when they are collected from different spatio-temporal scenarios and domains. Consequently, it is essential to enhance the model adaptation framework by enabling it to effectively bridge the distribution gap and capture spatio-temporal invariants. This enhancement will facilitate the successful knowledge transfer from the pre-trained phase to the downstream prediction tasks.

\noindent \textbf{Contributions}. To overcome the challenges mentioned above, we introduce a lightweight and innovative prompt-tuning framework (\model) that aims to achieve spatio-temporal in-context learning, enabling efficient and effective model adaptation across different spatio-temporal prediction tasks. Within our framework, we employ a spatio-temporal prompt network to enable in-context learning. To tackle challenge \textbf{C1}, we propose a context distillation mechanism that captures contextual signals from unseen data, facilitating adaptation to diverse spatio-temporal scenarios. Furthermore, a dependency modeling scheme is introduced to capture and understand the relationships between time and locations, enabling effective analysis of inter-dependencies among spatio-temporal elements. To overcome challenge \textbf{C2}, we enhance our \model\ framework by incorporating a unified distribution mapping mechanism, which bridges the distribution gap between pre-trained and downstream tasks. This mechanism facilitates the effective transfer of knowledge from the pre-trained insights to downstream spatio-temporal forecasting by aligning the data distributions through the regularization of prompt embeddings.

We conduct extensive experiments on four distinct types of spatio-temporal data tasks to evaluate the effectiveness of our proposed framework. The results demonstrate that our framework substantially improves the generalization capabilities in downstream prediction tasks across different spatio-temporal datasets. The noteworthy enhancement in predictive performance, achieved with model efficiency, underscores the efficacy of our \model\ framework. 


\begin{figure}
    \centering
    \subfigure{
        \begin{minipage}[t]{0.48\linewidth}
            \centering
            \includegraphics[width=1.53in]{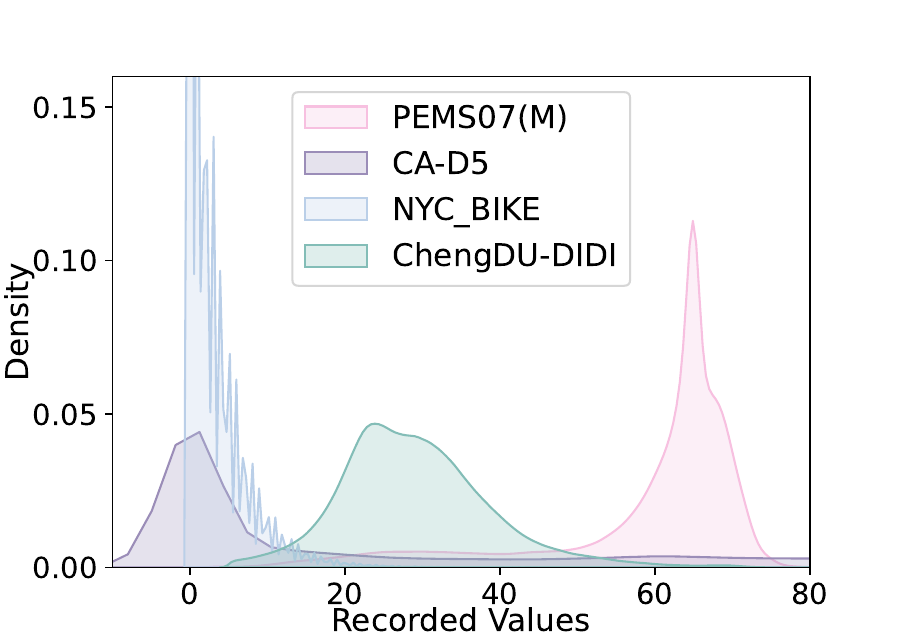}
        \end{minipage}%
        \begin{minipage}[t]{0.52\linewidth}
            \centering
            \includegraphics[width=1.60in]{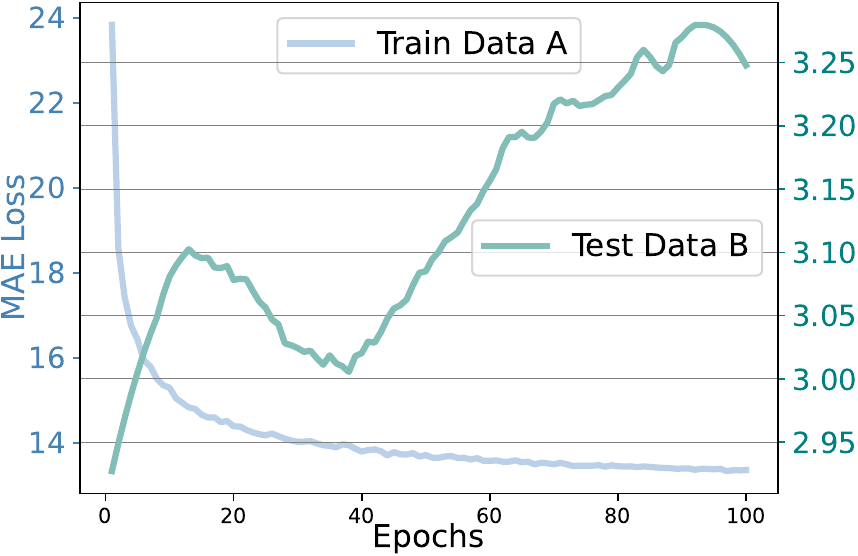}
        \end{minipage}%
    }%
    \centering
    \vspace{-0.15in}
    \caption{Motivations behind \model. The left figure illustrates the diverse data distributions across various ST datasets, while the right figure demonstrates that the end-to-end model's parameters are overfit to training set A and fail to generalize to test set B.}
    \vspace{-0.3in}
    \label{fig:intro}
\end{figure}
\section{Related Work}
\label{sec:relate}

\textbf{Deep Spatio-Temporal Learning} is a thriving research field that aims to model and understand the intricate spatio-temporal dynamics in real-world urban data. Various approaches have been proposed, including integrating recurrent neural networks (RNNs)~\cite{yao2018deep,Zhao2020TGCN} or Transformers\cite{wang2020traffic,xu2020spatial} to capture long-term patterns and short-term fluctuations. Attention mechanisms have been widely used to dynamically weigh spatial and temporal features~\cite{zheng2020gman,zhang2021traffic}. MLPs are also utilized as temporal encoders to capture temporal correlations~\cite{shao2022spatial}. Another important direction involves leveraging Graph Neural Networks (GNNs)~\cite{yu2018spatio,li2018diffusion,yu2018spatio,ye2021coupled} to simultaneously capture spatial dependencies and temporal evolutions. These models employ graph structures to represent relationships between spatial entities, facilitating information propagation. However, most existing deep spatio-temporal models adopt an end-to-end approach, limiting their generalization across diverse urban forecasting tasks and domains.

\textbf{Pre-Training with Spatio-Temporal Data.} There has been a recent surge of interest in using pre-training for acquiring comprehensive representations from spatio-temporal data through self-supervised learning. This approach aims to learn generalizable representations that capture the underlying patterns and dynamics. Two dominant research lines have emerged: i) integrating contrastive learning into the pre-training stage for spatio-temporal data~\cite{zhang2023spatial}, and ii) utilizing masked autoencoders~\cite{shao2022pretraining,li2023gpt} to reconstruct the masked input data to learn the spatio-temporal relationships.

\textbf{Prompt-Tuning} is a technique that optimizes prompts or instructions during inference to fine-tune the model, resulting in more accurate and context-specific predictions \cite{shrivastava2023repository,liu2023pre}. It has gained recognition as a promising approach to improve the performance and adaptability of pre-trained models in various domains. For example, in prompt-based language generation, it allows precise control over the generated output, aligning it with specific requirements or desired styles \cite{brown2020language,schick2020exploiting,zhang2023llama}. Similarly, in vision prompt learning, specific instructions guide the model's attention towards relevant visual features or concepts \cite{zhou2022learning,zhou2022conditional,sohn2023visual}. This study explores spatio-temporal prompt-tuning to enhance prediction models in urban computing.


\section{Preliminaries}
\label{sec:model}
\textbf{Representation of Spatial-Temporal Data}: The spatial-temporal information is captured and encoded using a three-way tensor $\textbf{X}$, where $\textbf{X}\in\mathbb{R}^{R\times T\times F}$. In this representation, $R$ represents the number of regions, $T$ denotes the time slots, and $F$ indicates the number of features. Each entry $\textbf{X}_{r,t,f}$ of the tensor corresponds to the $f$-th feature of the $r$-th region at the $t$-th time slot. For example, in the context of traffic flow prediction, the tensor $\textbf{X}$ captures traffic volume data, which quantifies the number of vehicles passing through a specific region within a fixed time interval (e.g., 5 minutes).

\textbf{Spatio-Temporal Prediction.}
In the context of spatio-temporal learning, a common scenario revolves around predicting future urban spatio-temporal conditions using historical records. This scenario can be described as follows:
\begin{align}
    \label{eq:Preliminarie1}
    \textbf{X}_{t_{K+1}: t_{K+P}}^{A} = g(\textbf{X}_{t_{K-H+1}: t_K}^{A})
\end{align}
In this scenario, $\textbf{X}^{A}$ represents the spatio-temporal data from dataset A. The function $g(\cdot)$ corresponds to the spatio-temporal predictive function, which could be methods like spatio-temporal transformers or spatio-temporal graph neural networks (GNNs). These predictive methods generate forecasts for the next $P$ time slots by leveraging the historical spatio-temporal data from the previous $H$ time slots.
\begin{align}
    \label{eq:Preliminarie2}
    \textbf{X}_{t_{K+1}: t_{K+P}}^{B} = \tilde{g}(f(\textbf{X}_{t_{K-H+1}: t_K}^{B}))
\end{align}
Furthermore, we introduce $\tilde{g}(\cdot)$ as the spatio-temporal prediction model with fixed parameters, and $\textbf{X}^{B}$ represents an unseen spatio-temporal dataset. The goal of this work is to enable swift adaptation of the $\tilde{g}(\cdot)$ model to a new dataset while ensuring parameter efficiency. To achieve this, we propose incorporating a lightweight and effective prompt-tuning paradigm into the pre-trained $\tilde{g}(\cdot)$ model.

\section{Methodology}
\label{sec:solution}
This section provides an in-depth explanation of the technical details of the proposed \model\ framework. We commence by outlining the foundational paradigm and design principles that guide the development of the \model\ framework. Subsequently, we present a comprehensive overview of the framework's main components: the Spatio-Temporal In-Context Learning and the Unified Distribution Mapping Mechanism. To provide a visual representation of the overall model architecture, please refer to Figure 2.

\begin{figure*}
    \centering
    \includegraphics[width=2.05\columnwidth]{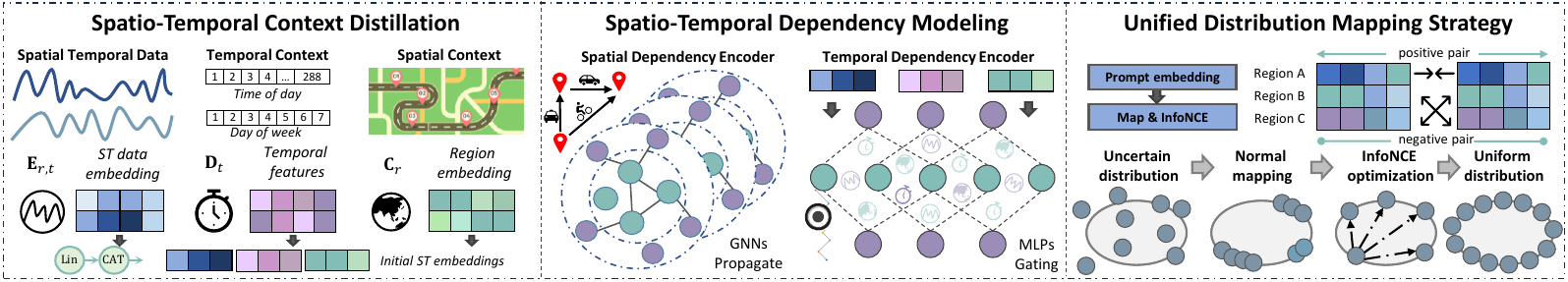}
    \vspace{-0.12in}
    \caption{Our proposed \model\ framework adopts an architecture that integrates spatio-temporal in-context learning and a unified distribution mapping mechanism, offering an efficient and effective approach for spatio-temporal prompt-tuning across diverse scenarios.}
    \vspace{-0.1in}
    \label{fig:fra01}
\end{figure*}

\subsection{Spatio-Temporal In-Context Learning}
\label{sec:prompt-net}
Our in-context learning framework is implemented through a spatio-temporal prompt network, which comprises two primary components: (i) Spatio-temporal context distillation mechanism efficiently captures contextual signals that are aware of both time and location from the unseen data. By doing so, it enables the model to learn from the specific context of the data, facilitating effective adaptation to various spatio-temporal scenarios. (ii) Spatio-temporal dependency modeling scheme incorporates complex relationships across time and locations into the in-context spatio-temporal network. By capturing and modeling these dependencies, the network can effectively understand the interdependencies and interactions between different spatio-temporal elements.

\subsubsection{\bf Spatio-Temporal Context Distillation}
\label{sec:st-embeddinhg}

\textbf{Spatio-Temporal Data Projection.} To initialize the representation of the spatio-temporal data, we employ a projection layer with two steps: normalization using the Z-Score function and augmentation through linear transformation:
\begin{align}
    \label{eq:initialize}
    \textbf{E}_{r,f} = \bar{\textbf{X}}_{r,f} \textbf{e}_0;~~~
    \bar{\textbf{X}}_{r,t,f} = \text{Z}(\textbf{X}_{r,t,f}) = \frac{\textbf{X}_{r,t,f}-\mu}{\sigma}
\end{align}
$\textbf{E}_{r,f}\in\mathbb{R}^d$ represents the representation for the $r$-th region of the $f$-th feature, where $d$ denotes the number of hidden units. $\bar{\textbf{X}}_{r,t,f}\in\mathbb{R}$ corresponds to the normalized spatio-temporal data feature obtained through the Z-Score ($\text{Z}(\cdot)$) function. The parameters $\mu$ and $\sigma$ represent the mean and standard deviation of the raw ST matrix $\textbf{X}$. The parametric vector $\textbf{e}_0\in\mathbb{R}^{T\times d}$ denotes the unit embedding vector for the normalized spatio-temporal feature.

\textbf{Temporal Context Incorporation.} In order to capture dynamic and periodic spatio-temporal patterns from diverse urban data, we introduce the time-aware context into our prompt network. This context is based on multi-resolution temporal features, specifically the hour of the day ($\textbf{z}^{(d)}\in\mathbb{R}^{T}$) and the day of the week ($\textbf{z}^{(w)}\in\mathbb{R}^{T}$). To effectively extract the temporal contextual signals, we employ a process of transformation and concatenation (denoted as $\text{CAT}[,]$), which can be summarized as follows:
\begin{align}
    \label{eq:temporal}
    \textbf{M}_t = \text{CAT}[\textbf{z}^{(d)}_t\cdot \textbf{e}_1,\textbf{z}^{(w)}_t\cdot \textbf{e}_2]
\end{align}

\textbf{Spatial Context Incorporation.}
To enhance the prompt network with geographical contextual information related to regional properties, we incorporate the citywide road network structure as an encoding feature that reflects spatial context. This process starts by formulating the normalized Laplacian matrix, which is defined as follows:
\begin{align}
    \label{eq:lpls}
    \bm{\triangle} = \bf{I - D^{-1/2}AD^{-1/2}}
\end{align}
Here, $\textbf{I}$, $\textbf{D}$, and $\textbf{A}$ denote the identity matrix, the degree matrix, and the adjacency matrix, respectively. The adjacency matrix is computed by considering the distances between areas and the road structure. Since Laplacian eigenvectors effectively preserve global graph structure information in Euclidean space, we perform eigenvalue decomposition to obtain $\bm{\triangle} = \bf{U\Lambda U^\top}$. After extracting the eigenvalue matrix $\bm{\Lambda}$ and the corresponding eigenvector matrix $\textbf{U}$, we derive the structure-aware node properties $\textbf{C}\in\mathbb{R}^{R\times d_r}$ by projecting $\textbf{U}$ to obtain the $d_r$ smallest nontrivial eigenvectors.

Acknowledging the potential disparity of $\textbf{C}$ in the feature space between training and test datasets, we employ a MLP to map these features and enhance the network's ability to generalize spatial context. Subsequently, we use the concatenation operation to integrate the aforementioned embeddings and obtain the initial spatio-temporal embedding:
\begin{align}
    \label{eq:node_eb}
    \bar{\textbf{E}} = \text{CAT}[\textbf{E},\textbf{M},\bar{\textbf{C}}];~~~
    \bar{\textbf{C}}_{r,f}=\text{MLP}(\textbf{C}_{r,f})
\end{align}
$\bar{\textbf{E}}$ denotes the integrated embeddings, where we concatenate the original representation $\textbf{E}$, the temporal context embeddings $\textbf{M}$, and the transformed regional properties $\bar{\textbf{C}}$. The transformation of $\textbf{C}$ is achieved through the MLP function.

\subsubsection{\bf Spatio-Temporal Dependency Modeling}
\label{sec:st-encoder}

\textbf{Temporal Dependency Encoder}. 
To capture dependencies across different time slots and preserve time-evolving data patterns, we introduce a lightweight gating mechanism. The formal operations of this mechanism are presented below:
\begin{align}
    \label{eq:temporal_encoder}
    \textbf{H}_{r,f} = (W_2 \sigma(W_1 \bar{\textbf{E}}_{r,f} +b_1) +b_2) + \bar{\textbf{E}}_{r,f}
\end{align} 
The trainable parameters of this gating mechanism are denoted by $W_1, W_2 \in \mathbb{R}^{d'\times d'}$ and $b_1, b_2 \in \mathbb{R}^{d'}$. The initial spatio-temporal embedding is represented by $\bar{\textbf{E}}_{r,f}$, while the encoded temporal embedding is denoted by $\textbf{H}_{r,f} \in \mathbb{R}^{d'\times d'}$. The embedding $\bar{\textbf{E}}_{r,f}$ contains valuable contextual information about temporal dynamics and regional characteristics, which is essential for the temporal dependency encoder. This enriched information enables our in-context learning to effectively identify variations in spatio-temporal patterns across different regions and time intervals, thereby facilitating precise modeling of temporal correlations.

\textbf{Spatial Dependency Encoder}.
Drawing inspiration from the effectiveness of graph neural networks in capturing spatial correlations between geographical locations~\cite{wu2020connecting,zheng2020gman}, we leverage graph convolution-based message passing to encode inter-regional correlations. The adjacency matrix $\textbf{A}$, as defined in Eq~{\ref{eq:lpls}}, acts as the connectivity matrix within the graph network framework. The process of spatial encoding is formalized as follows:
\begin{align}
    \label{eq:spatial_encoder}
    \textbf{S}_{f} = \sigma(\textbf{A} \cdot \textbf{H}_{f} \cdot W_3) + \textbf{H}_{f}
\end{align}
Here, $W_3\in\mathbb{R}^{d'\times d'}$ represents the trainable parameters. To mitigate the potential over-smoothing phenomenon caused by multi-layer GNNs, we employ residual networks. By stacking multiple layers of spatio-temporal encoders, our model generates a representation $\textbf{E}_{pro}$ that captures rich spatial and temporal dependencies across time and space.

\subsection{Unified Distribution Mapping Mechanism}
\label{sec:uniform}
To bridge the distribution gap between the pre-training and diverse unseen data in downstream tasks, we enhance \model\ by incorporating a distribution mapping mechanism. The objective of this mechanism is to transform both the pre-training data and the downstream data into a shared distribution space. This alignment of data distributions enables a seamless transfer of knowledge, ensuring that the insights gained from the pre-training stage can be effectively applied to the downstream spatio-temporal context.

To achieve this goal, our \model\ involves standardizing the prompt embedding to ensure a consistent distribution across diverse downstream datasets. We draw inspiration from various works in contrastive learning~\cite{wang2020understanding, chuang2020debiased, Wang2021understanding} and incorporate the infoNCE-based loss function to regularize the prompt network for representation generation. This loss function is designed to bring the representations of positive sample pairs closer together while pushing apart those of negative pairs. By leveraging self-supervised learning without the need for additional labeling, optimizing the infoNCE loss helps attain a more uniform embedding distribution. Empirical evidence suggests that through this loss alone, an almost entirely uniform distribution can be achieved~\cite{yu2022are}. Building upon this, we employ the infoNCE loss to adjust the distribution of the learned spatio-temporal prompt embedding $\textbf{E}_{pro}\in\mathbb{R}^{R\times F\times d'}$, formalized as:
\begin{align}
    \mathcal{L}_{Uni} = \sum_{r=1}^R\sum_{f=1}^F -\log{\sum_{r'} \exp(\cos(\textbf{E}_{pro}^{r,f}, \textbf{E}_{pro}^{r',f})/\tau)}
\end{align}
The cosine similarity function $\cos(\cdot)$ is used to measure the similarity between embeddings, and the temperature coefficient $\tau$ is used to adjust the softmax scale. 
In our \model, we enhance the uniformity of prompt embeddings by increasing the separation among embeddings that correspond to different regions. This improvement allows downstream models to effectively utilize the provided prompts $\textbf{E}^{(U)}_{pro}$ for rapid generalization across new data and tasks.

\subsection{Unifying Pre-training and Downstream Tasks}
Our spatio-temporal pretrain-prompt framework integrates the model pre-training process with downstream forecasting tasks through efficient prompt-tuning. In the pre-training phase, we train and optimize all parameters using dedicated pre-training datasets. Subsequently, in the prompt fine-tuning phase, we exclusively update the parameters of the prompt network by conducting a limited number of training epochs on unseen datasets. This approach enables the downstream model to effectively adapt to new data. Moreover, our framework is model-agnostic, allowing seamless integration with various existing spatio-temporal prediction baselines as the downstream model.


i) \textbf{Pre-training Phase}: Our goal is to forecast future trends using historical spatio-temporal records from the pre-training data. We achieve this with following process:
\begin{align}
    \label{eq:pretrain}
    \hat{\textbf{Y}}^{A} = g(\textbf{E}^{(U)}_{pro}); ~~~~\textbf{E}^{(U)}_{pro} = f_{Prompt}(\textbf{X}^{A})
\end{align}
Here, we have adjustable parameters for both $f_{Prompt}(\cdot)$ and $g(\cdot)$. To update these parameters, we utilize the loss function $\mathcal{L}$ in conjunction with the Adam optimizer. The loss function $\mathcal{L}$ is a combination of the regression loss $\mathcal{L}_r$ and the uniformity loss $\mathcal{L}_{Uni}$, with $\lambda$ as the balance coefficient between the two losses. The label is represented by $Y\in\mathbb{R}^{R\times T\times F}$. The regression loss $\mathcal{L}_r$ is defined as:
\begin{align}
    \label{eq:loss_r}
    \mathcal{L}_r = \frac{1}{RTF}\sum_{r=1}^R\sum_{t=1}^T\sum_{f=1}^F|(\textbf{Y}_{r,t,f} - \hat{\textbf{Y}}_{r,t,f})|
\end{align}
This formulation calculates the average absolute difference between the predicted values $\hat{\textbf{Y}}_{r,t,f}$ and the actual labels $\textbf{Y}_{r,t,f}$ across all regions $R$, time steps $T$, and features $F$.

ii) \textbf{Prompt-Tuning Phase}: We freeze the parameters of the downstream model and focus exclusively on fine-tuning the prompt network. This fine-tuning process is carried out on the test dataset to perform spatio-temporal prediction tasks, following the formulation described in Eq~\ref{eq:pretrain}:
\begin{align}
    \hat{\textbf{Y}}^{B} = \tilde{g}(f_{Prompt}(\textbf{X}^{B}))
\end{align}
Here, $\tilde{g}$ represents the downstream model with frozen parameters, and $B$ indicates the test datasets. During the prompt-tuning phase, we conduct a limited number of optimization epochs to generate high-quality spatio-temporal prompt representations for the downstream models.

\begin{table}[t]
    \centering
    \vspace{-0.1in}
    \caption{Statistics of pre-training dataset.}
    \label{tab:Dataset_pretrain}
    \scalebox{0.69}{
    \begin{tabular}{c|c|c|c|c}
        \hline
        Dataset & Data Record & \# Region & Time Steps & Sample Date\\
        \hline
        PEMS03 & traffic flow & 358 & 26208 & 1/Sep/2018 - 30/Nov/2018 \\
        PEMS04 & traffic flow & 307 & 16992 & 1/Jan/2018 - 28/Feb/2018\\
        PEMS07 & traffic flow & 883 & 26208 & 1/May/2017 - 30/Aug/2017\\
        PEMS08 & traffic flow & 170 & 17856 & 1/Jul/2016 - 31/Aug/2016\\
        \hline
    \end{tabular}
    }
    \vspace{-0.12in}
\end{table}

\begin{table}[t]
    \centering
    \vspace{-0.1in}
    \caption{Statistics of downstream tasks dataset.}
    \label{tab:Dataset_test}
    \scalebox{0.65}{
    \begin{tabular}{c|c|c|c|c}
        \hline
        Dataset & Data Record & \# Region & Time Steps & Sample Date\\
        \hline
        PEMS07(M) & traffic speed & 170 & 12672 & 1/Jul/2016 - 31/Aug/2016\\
        CA-D5 & traffic flow & 211 & 16992 & 1/Jan/2017 - 28/Feb/2017 \\
        ChengDu-DIDI & traffic index & 524 & 17280 & 1/Jan/2018 - 30/Apr/2018\\
        NYC Citi Bike & bike orders & 250 & 4368 & 1/Apr/2016 - 30/Jun/2016\\
        \hline
    \end{tabular}
    }
    \vspace{-0.12in}
\end{table}

\section{Evaluation}
\label{sec:eval}
In this section, we analyze the performance of our model by addressing the following several research questions:\vspace{-0.15in}
\begin{itemize}[leftmargin=*]
\item \textbf{RQ1}: Does \model\ effectively generalize pre-trained models to new spatio-temporal prediction data and tasks? \vspace{-0.25in}
\item \textbf{RQ2}: How does the efficiency of our model compare to that of end-to-end training and fine-tuning approaches? \vspace{-0.1in}
\item \textbf{RQ3}: What is the impact of the key components of our \model\ on the performance of downstream models? \vspace{-0.1in}
\item \textbf{RQ4}: How do parameters affect the model performance? \vspace{-0.15in}
\end{itemize}


\begin{table*}[t]
\renewcommand\arraystretch{1}
    \centering
    \caption{Overall performance on PEMS07(M), CA-D5, ChengDu-DIDI and NYC Citi Bike datasets in terms of \textit{MAE}, \textit{RMSE} and \textit{MAPE}.}
    \label{tab:overall_performance}
    \scalebox{0.88}{
    \begin{tabular}{c c |c c c | c c c | c c c | c c c}
        \hline
        \multirow{2}*{Model} & \multicolumn{1}{c}{Dataset} & \multicolumn{3}{c}{PEMS07(M)} &
        \multicolumn{3}{c}{CA-D5} & \multicolumn{3}{c}{ChengDu-DIDI} &
        \multicolumn{3}{c}{NYC Citi Bike}\\
        \cline{2-14}
        & \multicolumn{1}{c}{Metrics} & \multicolumn{1}{c}{MAE} & \multicolumn{1}{c}{RMSE} & \multicolumn{1}{c}{MAPE} & \multicolumn{1}{c}{MAE} & \multicolumn{1}{c}{RMSE} & \multicolumn{1}{c}{MAPE} & \multicolumn{1}{c}{MAE} & \multicolumn{1}{c}{RMSE} & \multicolumn{1}{c}{MAPE} & \multicolumn{1}{c}{MAE} & \multicolumn{1}{c}{RMSE} & \multicolumn{1}{c}{MAPE}\\
        \cline{1-14}
        \multicolumn{2}{c}{TGCN} & 5.05 & 8.56 & 13.68\% & 15.36 & 24.52 & 24.42\% & 3.18 & 4.72 & 14.48\% & 2.19 & 3.54 & 63.31\% \\
        \multicolumn{2}{c}{STGCN} & 3.14 & 6.25 & 8.05\% & 15.02 & 23.73 & 27.84\% & 2.59 & 3.97 & 12.20\% & 2.04 & 3.21 & 52.06\% \\
        \multicolumn{2}{c}{ASTGCN} & 3.13 & 6.19 & 8.09\% & 15.26 & 24.37 & 24.45\% & 2.56 & 3.83 & 11.58\% & 2.03 & 3.10 & 55.13\% \\
        \multicolumn{2}{c}{GWN} & 2.67 & 5.40 & 6.62\% & 13.63 & 22.30 & 24.68\% & 2.35 & 3.56 & 10.72\% & 1.81 & 2.71 & 50.05\% \\
        \multicolumn{2}{c}{STSGCN} & 2.93 & 5.87 & 7.52\% & 15.30 & 24.44 & 24.70\% & 2.86 & 4.29 & 13.02\% & 1.89 & 2.78 & 52.48\% \\
        \multicolumn{2}{c}{AGCRN} & 2.89 & 5.69 & 7.32\% & 14.44 & 23.10 & 23.02\% & 2.87 & 4.27 & 13.05\% & 2.16 & 3.41 & 60.95\% \\
        \multicolumn{2}{c}{MTGNN} & 2.70 & 5.50 & 6.81\% & 13.90 & 22.44 & 23.86\% & 2.33 & 3.56 & 10.63\% & 1.84 & 2.80 & \textbf{48.69\%} \\
        \multicolumn{2}{c}{STFGNN} & 3.14 & 6.21 & 7.81\% & 15.38 & 24.37 & 25.78\% & 2.83 & 4.24 & 13.05\% & 2.36 & 3.73 & 61.93\% \\
        \multicolumn{2}{c}{STGODE} & 2.81 & 5.57 & 7.09\% & 14.13 & 22.77 & 23.41\% & 2.51 & 3.75 & 11.48\% & 1.94 & 2.83 & 56.25\% \\
        \multicolumn{2}{c}{DMSTGCN} & 2.94 & 5.92 & 7.45\% & 13.88 & 22.81 & 21.97\% & 2.67 & 4.00 & 12.34\% & 2.09 & 3.32 & 60.00\% \\
        \multicolumn{2}{c}{MSDR} & 2.79 & 5.53 & 7.07\% & 15.31 & 23.71 & 33.01\%  & 2.56 & 3.85 & 11.94\% & 1.87 & 2.86 & 51.25\% \\
        \multicolumn{2}{c}{STWA} & 2.74 & 5.52 & 6.97\% & 14.00 & 22.75 & 23.22\% & 2.44 & 3.68 & 11.33\% & 1.88 & 2.83 & 49.93\% \\
        \multicolumn{2}{c}{PDFormer} & 2.68 & 5.46 & 6.66\% & 13.85 & 22.59 & 23.11\% & 2.43 & 3.70 & 11.17\% & 1.89 & 2.88 & 51.98\% \\
        \hline
        \hline
        \multicolumn{2}{c}{Ours} & \textbf{2.59} & \textbf{5.25} & \textbf{6.54\%} & \textbf{13.26} & \textbf{21.83} & \textbf{21.49\%} & \textbf{2.31} & \textbf{3.52} & \textbf{10.51\%} & \textbf{1.79} & \textbf{2.63} & 49.77\% \\
        \cline{1-14}
    \end{tabular}
    }
    \vspace{-0.05in}
\end{table*}

\subsection{Experimental Setting}\vspace{-0.1in}

\textbf{Datasets}. To evaluate how well the model can generalize across different urban spatio-temporal contexts, our experimental setup is designed as follows: (i) In the pre-training phase, we use four datasets (PEMS03, PEMS04, PEMS07, and PEMS08~\cite{Song2020spatial}) as our training sets. These datasets consist of records detailing traffic flow conditions across various streets and cities in California, USA. (ii) During the subsequent prompt-tuning phase, we concentrate on four distinct target datasets to fine-tune and evaluate our framework: PEMS07(M)~\cite{yu2018spatio}, CA-D5~\cite{liu2023largest}, ChengDu-DIDI~\cite{lu2022spatio}, and NYC Citi Bike~\cite{ye2021coupled}. These datasets correspondingly represent traffic speeds in Los Angeles, traffic flow in California, traffic flow in Chengdu, and bicycle demand in New York City. Each target dataset is partitioned into training, validation, and test sets in a ratio of 6:2:2. Detailed statistics for each dataset can be found in Table~\ref{tab:Dataset_pretrain} and Table~\ref{tab:Dataset_test}.



\textbf{Experimental Description. } 
We configure the model with 32 hidden units for $d$, $d_t$, and $d_r$, while employing a spatial and temporal encoder with 2 layers. The pre-training phase involves alternating training for 300 epochs on the four pre-training datasets. Subsequently, we conduct fine-tuning for 20 epochs during the prompt-tuning phase. In both the baseline and full fine-tuning setups, the maximum number of training epochs is limited to 100. Furthermore, a batch size of 64 is employed for both phases and all baselines. For the dataset-specific parameters in the baselines (\eg, node embeddings), we initialize them randomly and make them trainable during the prompt fine-tuning phase. Additionally, the final regression layers of some baseline models are set to be trainable to ensure accurate numerical predictions.

\textbf{Evaluation Metrics.} In order to assess the performance of the model in spatiotemporal prediction, we employ three widely-used evaluation metrics: Mean Absolute Error (MAE), Root Mean Squared Error (RMSE), and Mean Absolute Percentage Error (MAPE). These metrics quantify the disparities between the predicted data and the ground truth data. It is important to note that lower values of these metrics indicate better performance.


\textbf{Baseline Description.} To evaluate the effectiveness of our \model, we selected 13 advanced spatio-temporal prediction models as baselines. These include RNNs-based models, attention-based models, GNNs-based models, and differential equation-based spatio-temporal prediction methods.


\noindent \textbf{RNN-based Spatio-Temporal Prediction Approaches:}
\begin{itemize}[leftmargin=*]
\item \textbf{AGCRN}~\cite{bai2020adaptive}: This method incorporates RNNs with learnable node embeddings to capture personalized spatio-temporal patterns of regions.
\item \textbf{MSDR}~\cite{liu2022msdr}: This model addresses the issue of long-term information forgetting in RNNs by introducing a variant of RNNs. It further combines this variant with GNNs to capture spatio-temporal correlations.
\end{itemize}

\textbf{GNN-based Spatio-Temporal Prediction Models:}
\begin{itemize}[leftmargin=*]
\item \textbf{TGCN}~\cite{Zhao2020TGCN}: This model leverages a combination of RNNs and GCNs to separately model temporal dependencies and spatial correlations.
\item \textbf{STGCN}~\cite{yu2018spatio}: This method utilizes gated convolutional networks to encode temporal dependencies and GCNs to capture local spatial relationships.
\item \textbf{GWN}~\cite{wu2019graph}: This approach incorporates a learnable graph structure to model spatial correlations, while TCNs are used to extract temporal features.
\item \textbf{STSGCN}~\cite{Song2020spatial}: It synchronously models dynamic spatio-temporal correlations by constructing a graph network that allows interactions across time slices.
\item \textbf{MTGNN}~\cite{wu2020connecting}: This method captures temporal dependencies using time convolutional networks combined with skip connections, and employs learnable graph networks to model spatial correlations.
\item \textbf{STFGNN}~\cite{li2021spatial}: This method proposes a data-driven spatio-temporal graph construction approach to capture dynamic spatio-temporal correlations.
\item \textbf{DMSTGCN}~\cite{han2021dynamic}: This method models dynamic spatial correlations by constructing a learnable dynamic graph network.
\end{itemize}

\noindent \textbf{Attention-based Spatio-Temporal Models:}
\begin{itemize}[leftmargin=*]
\item \textbf{ASTGCN}~\cite{Guo2019attention}: This method leverages attention mechanisms to effectively capture the periodic spatio-temporal correlations inherent in the data.
\item \textbf{STWA}\cite{cirstea2022towards}: This approach goes beyond standard attention mechanisms by incorporating specialized node features and time-dynamic parameters.
\item \textbf{PDFormer}~\cite{jiang2023pdformer}: This model leverages the power of the transformer architecture to effectively encode the intricate spatio-temporal dependencies present in the data. Furthermore, it introduces a novel time-delayed spatial correlation modeling technique.
\end{itemize}

\textbf{ODE-based Spatio-Temporal Prediction Methods:}
\begin{itemize}[leftmargin=*]
\item \textbf{STGODE}~\cite{fang2021spatial}: This method enriches graph networks with ordinary differential equations (ODEs) to extract continuous spatial correlation information. Time convolutional networks are employed to learn time-dependent relationships within the data.
\end{itemize}

\begin{table*}[htbp]
\renewcommand\arraystretch{1}
    \centering
    \caption{Model-agnostic experiments on PEMS07(M), CA-D5, ChengDu-DIDI and NYC Citi Bike datasets.}
    \label{tab:model-ag}
    \scalebox{0.87}{
    \begin{tabular}{c c |c c c | c c c | c c c | c c c}
        \hline
        \multirow{2}*{Model} & \multicolumn{1}{c}{Dataset} & \multicolumn{3}{c}{PEMS07(M)} &
        \multicolumn{3}{c}{CA-D5} & \multicolumn{3}{c}{ChengDu-DIDI} &
        \multicolumn{3}{c}{NYC Citi Bike}\\
        \cline{2-14}
        & \multicolumn{1}{c}{Metrics} & \multicolumn{1}{c}{MAE} & \multicolumn{1}{c}{RMSE} & \multicolumn{1}{c}{MAPE} & \multicolumn{1}{c}{MAE} & \multicolumn{1}{c}{RMSE} & \multicolumn{1}{c}{MAPE} & \multicolumn{1}{c}{MAE} & \multicolumn{1}{c}{RMSE} & \multicolumn{1}{c}{MAPE} & \multicolumn{1}{c}{MAE} & \multicolumn{1}{c}{RMSE} & \multicolumn{1}{c}{MAPE}\\
        \cline{1-14}

        \multicolumn{2}{c}{STGCN} & 3.14 & 6.25 & 8.05\% & 15.02 & 23.73 & 27.84\% & 2.59 & 3.97 & 12.20\% & 2.04 & 3.21 & 52.06\% \\
        \multicolumn{2}{c}{w/o Finetune} & 8.07 & 11.77 & 23.02\% & 31.20 & 41.82 & 68.75\% & 4.82 & 6.53 & 22.87\% & 4.10 & 5.60 & 148.20\% \\
        \multicolumn{2}{c}{w/ Finetune} & 3.18 & 6.25 & 8.04\% & 15.15 & 24.18 & 23.30\% & 2.58 & 3.93 & 12.01\% & 1.99 & 3.11 & 50.38\% \\
        \multicolumn{2}{c}{Ours} & \textbf{2.68} & \textbf{5.37} & \textbf{6.80\%} & \textbf{13.47} & \textbf{22.01} & \textbf{22.62\%} & \textbf{2.37} & \textbf{3.60} & \textbf{10.92\%} & \textbf{1.80} & \textbf{2.67} & \textbf{50.06\%} \\
        \cline{1-14}
        \multicolumn{2}{c}{GWN} & 2.67 & 5.40 & \textbf{6.62\%} & 13.63 & 22.30 & 24.68\% & 2.35 & 3.56 & 10.72\% & 1.81 & 2.71 & \textbf{50.05\%} \\
        \multicolumn{2}{c}{w/o Finetune} & 7.41 & 12.78 & 24.30\% & 27.28 & 41.46 & 48.26\% & 4.89 & 6.99 & 23.42\% & 3.39 & 4.99 & 112.57\% \\
        \multicolumn{2}{c}{w/ Finetune} & 2.69 & 5.38 & 6.74\% & 13.61 & 22.44 & 22.82\% & \textbf{2.35} & \textbf{3.55} & \textbf{10.72\%} & 1.85 & 2.79 & 51.88\% \\
        \multicolumn{2}{c}{Ours} & \textbf{2.67} & \textbf{5.36} & 6.79\% & \textbf{13.19} & \textbf{21.68} & \textbf{21.21\%} & 2.37 & 3.58 & 10.77\% & \textbf{1.80} & \textbf{2.66} & 51.16\% \\
        \cline{1-14}
        \multicolumn{2}{c}{MTGNN} & 2.70 & 5.50 & 6.81\% & 13.90 & 22.44 & 23.86\% & 2.33 & 3.56 & 10.63\% & 1.84 & 2.80 & \textbf{48.69\%} \\
        \multicolumn{2}{c}{w/o Finetune} & 5.91 & 9.21 & 14.56\% & 50.07 & 65.35 & 77.26\% & 4.45 & 6.27 & 20.06\% & 3.32 & 4.74 & 118.01\% \\
        \multicolumn{2}{c}{w/ Finetune} & 2.62 & 5.33 & 6.56\% & 13.46 & 22.02 & 22.10\% & 2.33 & 3.54 & 10.52\% & 1.81 & 2.74 & 49.35\% \\
        \multicolumn{2}{c}{Ours} & \textbf{2.59} & \textbf{5.25} & \textbf{6.54\%} & \textbf{13.26} & \textbf{21.83} & \textbf{21.49\%} & \textbf{2.31} & \textbf{3.52} & \textbf{10.51\%} & \textbf{1.79} & \textbf{2.63} & 49.77\% \\
        \cline{1-14}
        \multicolumn{2}{c}{PDFormer} & 2.68 & 5.46 & \textbf{6.66\%} & 13.85 & 22.59 & 23.11\% & 2.43 & 3.70 & 11.17\% & 1.89 & 2.88 & 51.98\% \\
        \multicolumn{2}{c}{w/o Finetune} & 5.57 & 10.03 & 19.16\% & 20.45 & 31.56 & 30.21\% & 3.60 & 5.20 & 16.93\% & 3.19 & 4.76 & 111.40\% \\
        \multicolumn{2}{c}{w/ Finetune} & 2.71 & 5.44 & 6.80\% & 13.77 & 22.31 & 24.90\% & 2.43 & 3.69 & 11.15\% & 1.83 & 2.78 & \textbf{48.56\%} \\
        \multicolumn{2}{c}{Ours} & \textbf{2.68} & \textbf{5.31} & 6.79\% & \textbf{13.55} & \textbf{22.08} & \textbf{21.98\%} & \textbf{2.41} & \textbf{3.64} & \textbf{11.05\%} & \textbf{1.81} & \textbf{2.68} & 52.06\% \\
        \cline{1-14}
    \end{tabular}
    }
    \vspace{-0.05in}
\end{table*}

\subsection{Overall Performance (RQ1)}
In this section, we investigate the effectiveness of \model\ in enhancing the generalization capabilities of downstream models across new spatio-temporal contexts. As shown in Table~\ref{tab:overall_performance}, the baseline models were exclusively trained and tested on the target dataset to evaluate their performance.


\textbf{Superiority over End-to-End Prediction Models.} Upon analyzing the results presented in Table~\ref{tab:overall_performance}, it becomes evident that our approach exhibits a significant advantage over the end-to-end spatio-temporal models in diverse urban data prediction scenarios. These findings provide compelling evidence of the effectiveness of our \model\ method in accurately capturing the intricate spatio-temporal invariant patterns present in urban data. Our new in-context learning paradigm excels in transferring this acquired knowledge to adapt the pre-trained model to new downstream tasks. By effectively handling the distribution gap, our method bridges the semantic gap between the pre-trained model and the specific prediction scenarios encountered in practice.



\textbf{Model-Agnostic Advantage.} The model-agnostic advantage is a notable strength of our approach. We have designed our method to seamlessly integrate with various spatio-temporal backbone encoders, providing flexibility and avoiding the constraints of a specific model choice. This adaptability is demonstrated in Table~\ref{tab:model-ag}, where our \model\ approach showcases easy adaptation with four state-of-the-art spatio-temporal models (i.e., STGCN, GWN, MTGNN, PDFormer). The evaluation results highlight the versatility of our \model\ approach, showcasing its exceptional performance enhancement when combined with these cutting-edge spatio-temporal models. The successful integration of our approach with state-of-the-art models further reinforces its adaptability and capability to improve prediction accuracy in diverse urban data scenarios.


\textbf{Comparison with Model Fine-Tuning.} To further demonstrate the effectiveness of our spatio-temporal in-context framework, we compare our prompt-tuning approach with the full fine-tuning of the model parameters. The "w/o Finetune" approach refers to direct prediction on the target dataset without any fine-tuning after pre-training. On the other hand, "w/ Finetune" indicates the utilization of full-parameter fine-tuning to adapt the models to the target data after pre-training. However, it is worth noting that the observed improvements with direct fine-tuning, compared to end-to-end prediction outcomes, suggest that pre-training may not have provided a significant advantage. In the absence of effective alignment between the pre-trained model and downstream tasks, noise can be introduced, leading to misleading fine-tuning and suboptimal performance.

\subsection{Assessment of Model Efficiency (RQ2)}\vspace{-0.05in}
\noindent \textbf{Training Time}. In this section, we focus on evaluating the efficiency of our proposed method. We measured the training time for three different scenarios: end-to-end training, full fine-tuning, and our \model\ approach. The results are presented in Table~\ref{tab:training_time}. For end-to-end training and full fine-tuning, we followed the settings of existing baselines, configuring the training epochs to 100. We also set an early stopping criterion of 25, which halts the training process if no decrease in validation loss is observed. Regarding our \model\ approach, we limited the number of epochs for prompt-tuning to 20. This constraint was implemented to promote swift adaptation of the downstream models to the new datasets. The results suggest that the same baseline model showcases comparable efficiency in both end-to-end training and full-parameter fine-tuning. The disparity in training time between these two settings primarily arises from variations in convergence speed caused by different initialization parameters. One of the significant contributions of our \model\ framework is the notable increase in computational efficiency. It reduces the training time of baseline models by 20\% to 80\%. This substantial enhancement significantly improves their efficiency in adapting to new spatio-temporal data.

\begin{figure}
\centering
\subfigure{
    \begin{minipage}[t]{0.5\linewidth}
        \centering
        \includegraphics[width=1.55in]{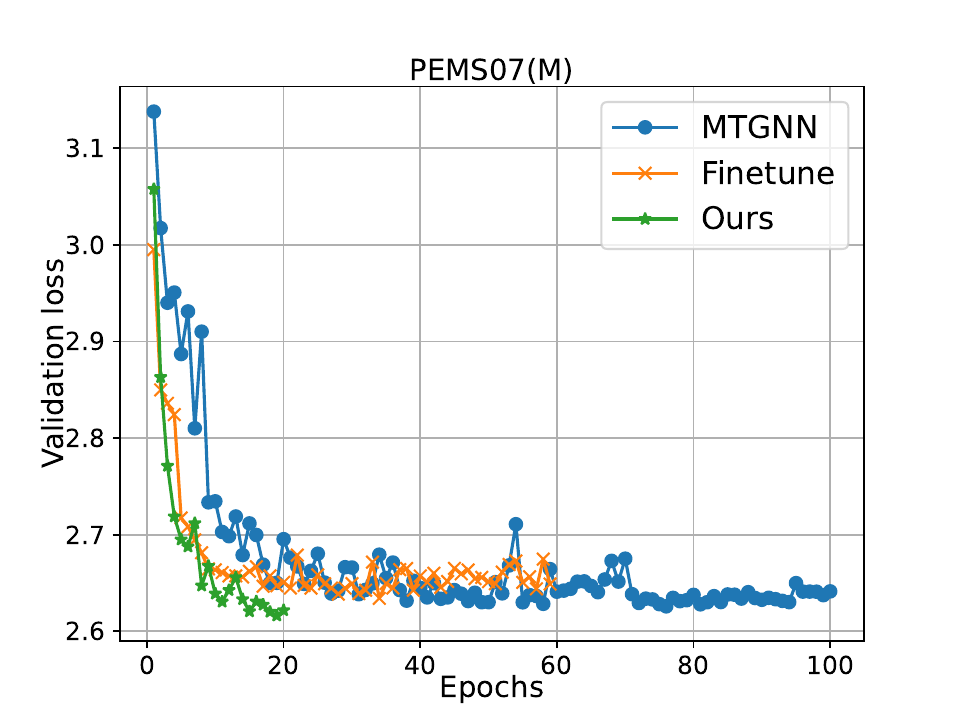}
    \end{minipage}%
    \begin{minipage}[t]{0.5\linewidth}
        \centering
        \includegraphics[width=1.55in]{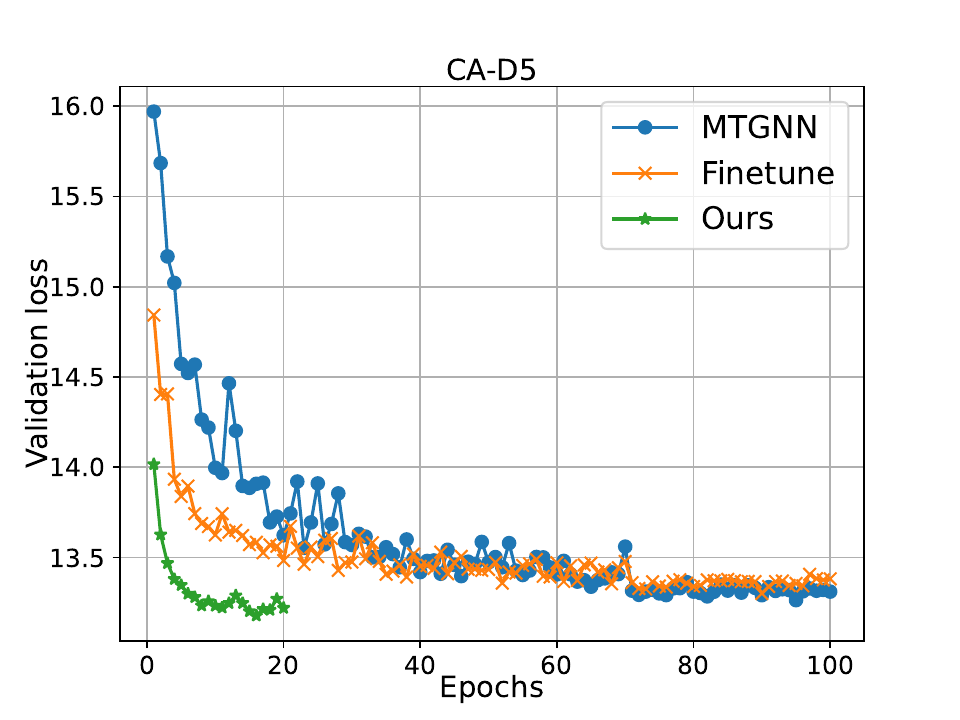}
    \end{minipage}%
}%

\centering
\vspace{-0.2cm}
\caption{The convergence efficiency of \model.}
\vspace{-0.1cm}
\label{fig:valloss}
\end{figure}

\textbf{Faster Convergence Rate.} To further validate the ability of our \model\ framework to swiftly adapt to new spatio-temporal scenarios, we conducted an investigation into its convergence speed on different datasets. We specifically focused on the decreasing trend of the validation error when using the PEMS07(M) and CA-D5 datasets, with MTGNN adopted as the downstream model. The results are visualized in Figure~\ref{fig:valloss}. The findings clearly demonstrate that with the integration of our \model\ approach, the downstream model achieves convergence within a few tuning epochs. In contrast, the end-to-end training and fine-tuning paradigms require a greater number of training rounds to fit the new data. This phenomenon can be attributed to the effectiveness of our proposed spatio-temporal prompt network and data mapping strategy. These components enable the model to leverage the spatio-temporal characteristics of the new data in conjunction with the pre-trained knowledge, facilitating rapid adaptation to diverse spatio-temporal scenes.

\begin{table}[t]
    \centering
    \vspace{-0.05in}
    \caption{Computational time cost investigation (seconds).}
    \label{tab:training_time}
    \scalebox{0.9}{
    \begin{tabular}{cr||cr}
        \hline
        Model & time(s) & Model & time(s)\\
        \hline
        ASTGCN & 1310 & TGCN & 927 \\
        AGCRN & 689 & STSGCN & 2253 \\
        DMSTGCN & 1330 & STFGNN & 990 \\
        MSDR & 5190 & STWA & 5143 \\
        \hline
        STGCN & 411 & GWN & 1042\\
        STGCN Finetune & 445 & GWN Finetune & 1185\\
        STGCN w/ Ours & 190 & GWN w/ Ours & 222\\
        \cline{1-4}
        MTGNN & 962 & PDFormer & 7524\\
        MTGNN Finetune & 1095 & PDFormer Finetune & 6466\\
        MTGNN w/ Ours & 230 & PDFormer w/ Ours & 1220\\
        \hline
    \end{tabular}}
\end{table}

\subsection{Ablation Study (RQ3)}
\begin{figure}
\centering
\subfigure{
    \begin{minipage}[t]{1\linewidth}
        \centering
        \includegraphics[width=3.19in]{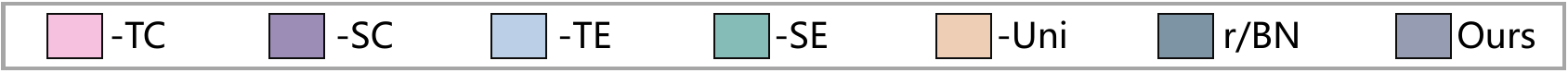}
    \end{minipage}%
}%
\vspace{-0.05in}
\subfigure[MAE, RMSE and MAPE on PEMS07(M) dataset.]{
    \vspace{-0.08in}
    \begin{minipage}[t]{0.333\linewidth}
        \centering
        \includegraphics[width=1.01in]{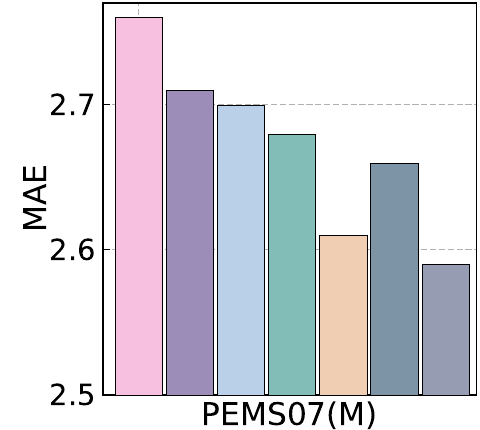}
    \end{minipage}%
    \begin{minipage}[t]{0.333\linewidth}
        \centering
        \includegraphics[width=1.01in]{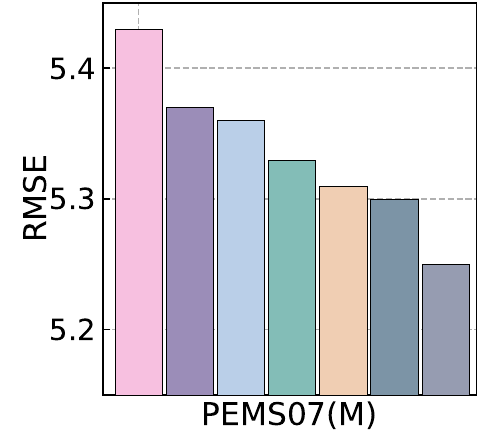}
    \end{minipage}%
    \begin{minipage}[t]{0.333\linewidth}
        \centering
        \includegraphics[width=1.01in]{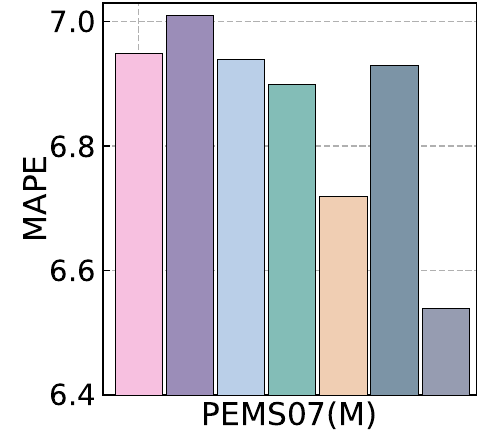}
    \end{minipage}%
}%
\vspace{-0.1in}
\subfigure[MAE, RMSE and MAPE on CA-D5 dataset.]{
    \vspace{-0.08in}
    \begin{minipage}[t]{0.335\linewidth}
        \centering
        \includegraphics[width=1in]{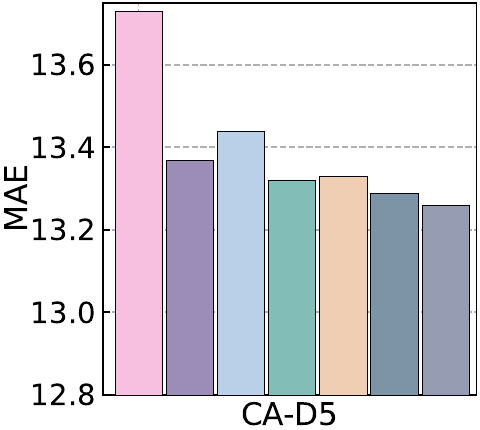}
    \end{minipage}%
    \begin{minipage}[t]{0.335\linewidth}
        \centering
        \includegraphics[width=1in]{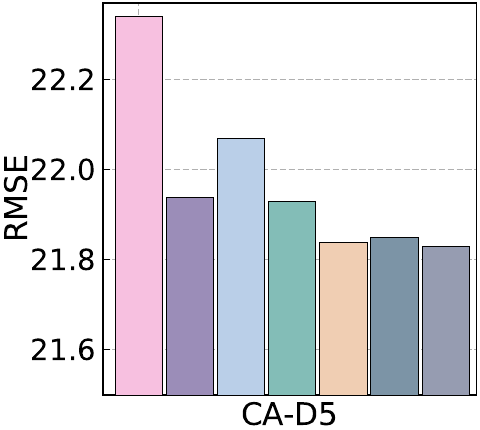}
    \end{minipage}%
    \begin{minipage}[t]{0.335\linewidth}
        \centering
        \includegraphics[width=1in]{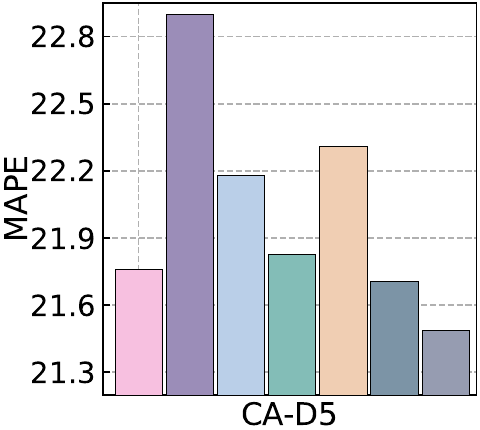}
    \end{minipage}%
}%

\centering
\vspace{-0.4cm}
\caption{Ablation study of \model.}
\vspace{-0.35cm}
\label{fig:ablation}
\end{figure}

\vspace{-0.05in}
In this section, we conduct an ablation study to investigate the contributions of each key module in our \model\ framework. The corresponding experimental results are depicted in Figure~\ref{fig:ablation}. We utilize MTGNN as the downstream baseline and assess the performance of each component on the PEMS07(M) and CA-D5 data. Before the prompt-tuning step, we conduct re-pre-training to adapt to each variant. \vspace{-0.1in}

\begin{itemize}[leftmargin=*]
\item \textbf{Impact of Spatio-Temporal Context Distillation.} 
To assess the importance of temporal and spatial context distillation, we conducted experiments by removing the distillation with temporal context (-TC) and spatial context (-SC) separately.  The results clearly demonstrate a substantial performance drop across most metrics when the distilled temporal and spatial context are removed. This highlights the critical importance of preserving both temporal and spatial context during the in-context learning process. 
Effective encoding of the temporal information and the incorporation of the spatial information are vital for capturing dynamic periodic temporal patterns and region-specific properties, thereby enhancing the model's ability to recognize spatio-temporal invariant patterns and improve its data understanding.
\vspace{-0.05in}

\item \textbf{Impact of Spatio-Temporal Dependency Modeling.} 
In our analysis, we individually removed the temporal encoder (-TE) and spatial encoder (-SE) to investigate their contributions. Our analysis revealed that spatio-temporal dependency encoding plays a vital role in effectively integrating complex relationships among different time slots and locations during the in-context learning process. The inclusion of both temporal and spatial dependency encoders enables the model to comprehend and leverage the intricate interactions between time and space. This proficiency facilitates a swifter adaptation of downstream models to new spatio-temporal scenarios. \vspace{-0.05in}

\item \textbf{Impact of Unified Distribution Mapping Mechanism.} 
In our variants, we assess the utility of the unified distribution mapping strategy through two aspects.
\textbf{i) -Uni.} We did not explore the unified distribution mapping strategy. However, the decline in performance indicates its positive impact on the model. By mapping the diverse spatio-temporal data embeddings into a uniform distribution, our \model\ effectively mitigates the influence of distributional disparities between the pre-training data and unseen spatio-temporal data. 
\textbf{ii) r/BN.} The unified distribution mapping strategy has been replaced with batch normalization. Batch normalization standardizes the data based on local statistical properties of mini-batches, mitigating the issue of internal covariate shift during neural network training and thereby enhancing convergence efficiency. However, due to the lack of an established connection between the pretrained data and the downstream task data, it is challenging for the downstream model to effectively transfer knowledge from the pretraining process.
The unified distribution mapping strategy ensures that the model can effectively leverage the knowledge acquired during the pre-training. By aligning the distributions of different data sources, the model can better adapt to new spatio-temporal scenarios and make more accurate predictions.

\vspace{-0.05in}

\end{itemize}



\subsection{Hyperparameter Analysis (RQ4)} \vspace{-0.05in}
In this section, we investigate the impact of different hyperparameter settings, specifically across various temperature coefficients and loss weight coefficients, on the model's performance. Our findings reveal that the model achieves optimal performance with a parameter configuration of $\tau=0.3$ and $\lambda=1.0$, as shown in Figure~\ref{fig:case}. Notably, variations in these parameters minimally influence the final result, highlighting the model's effective adaptability to different parameters settings. It learns efficient representations that distinguish embedded features in various regions, even with differences in feature scales. Additionally, the model's performance remains unaffected by an increase in the uniformity loss. This indicates that the key component of our distribution mapping strategy does not interfere with the predictive loss. This further supports the viability of our strategy and facilitates the downstream model's rapid generalization to novel spatio-temporal contexts.

\subsection{Case Study (RQ5)}\vspace{-0.05in}
To assess the effectiveness of our proposed uniform distribution mapping approach in transforming various data representations into uniform distributions, we conducted visualizations of prompt embeddings with and without the application of our distribution mapping mechanism. These visualizations were performed using two variants of the downstream model MTGNN and evaluated on the test sets of the PEMS07(M) and CA-D5 datasets. Given that prompt embeddings are high-dimensional vectors, we initially applied the PCA technique to reduce the dimensionality of each embedding sample to 2-D~\cite{svante1987principal}. Subsequently, we projected the reduced embeddings onto the unit circle using the L2 norm, as depicted in Figure~\ref{fig:case}. The visualization results provide compelling evidence that our uniform distribution mapping strategy effectively transforms the prompt embeddings into an approximate uniform distribution. In contrast, the variants that lack this strategy fail to achieve such desirable distribution properties. By converting data from new spatio-temporal contexts into a consistent distribution, our \model\ gains the ability to utilize pre-trained knowledge and rapidly adapt to new datasets, facilitating its performance on various spatio-temporal tasks.

\begin{figure}
\centering
\vspace{-0.11in}
\subfigure{
    \begin{minipage}[t]{0.5\linewidth}
        \centering
        \includegraphics[width=1.75in]{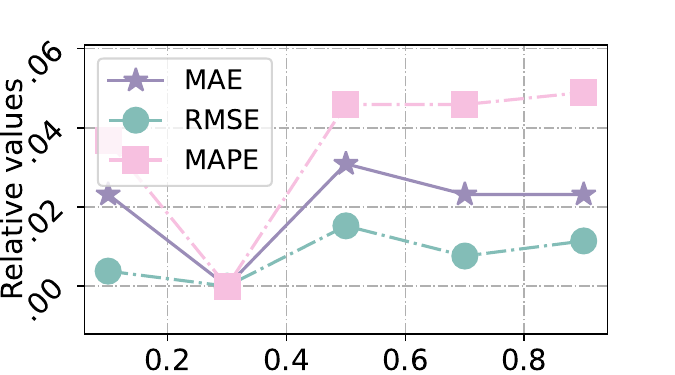}
    \end{minipage}%
    \begin{minipage}[t]{0.5\linewidth}
        \centering
        \includegraphics[width=1.75in]{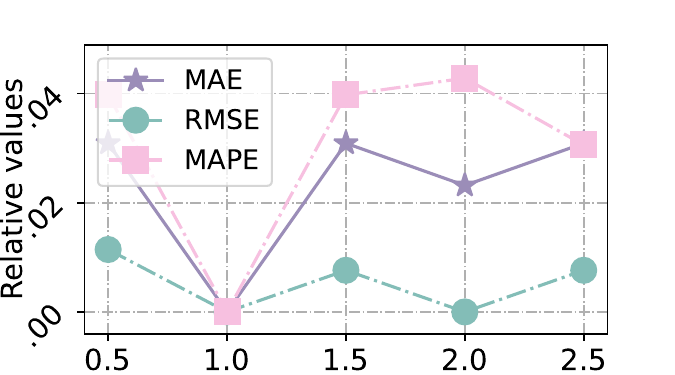}
    \end{minipage}%
}%

\centering
\vspace{-0.35cm}
\caption{Hyperparameter study of $\tau$ and $\lambda$.}
\vspace{-0.2cm}
\label{fig:hyperpara}
\end{figure}

\begin{figure}
\centering
\subfigure{
    \begin{minipage}[t]{0.25\linewidth}
        \centering
        \includegraphics[width=0.79in]{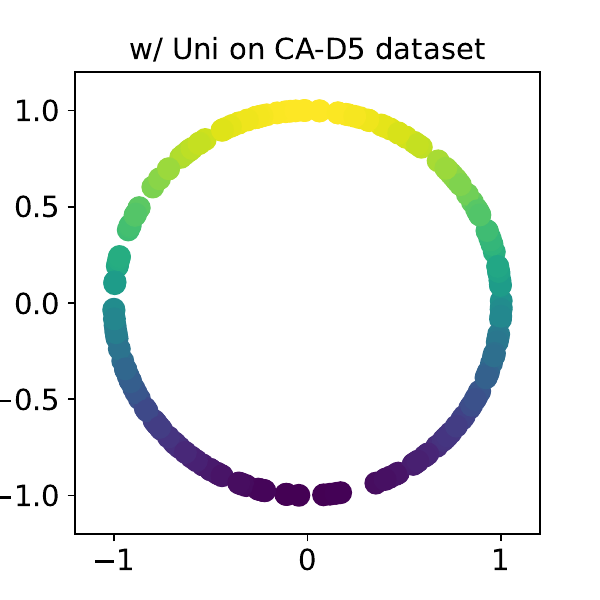}
    \end{minipage}%
    \begin{minipage}[t]{0.25\linewidth}
        \centering
        \includegraphics[width=0.79in]{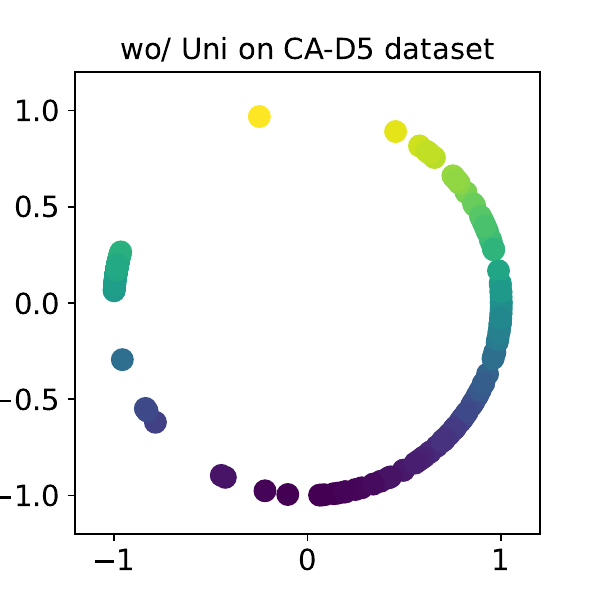}
    \end{minipage}%
    \begin{minipage}[t]{0.25\linewidth}
        \centering
        \includegraphics[width=0.79in]{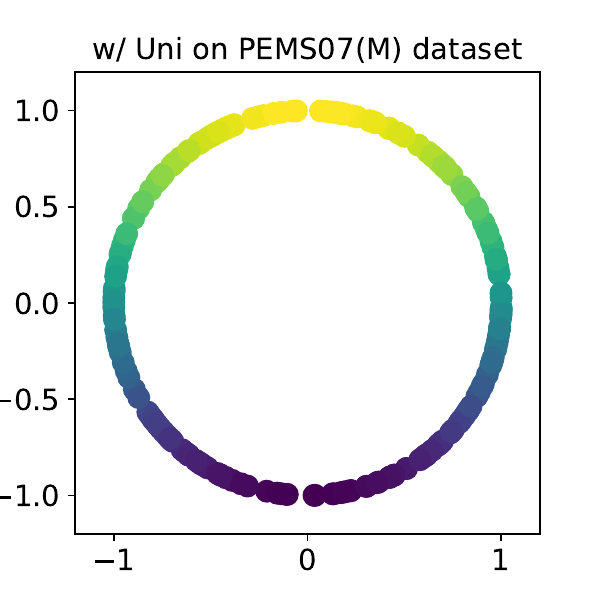}
    \end{minipage}%
    \begin{minipage}[t]{0.25\linewidth}
        \centering
        \includegraphics[width=0.79in]{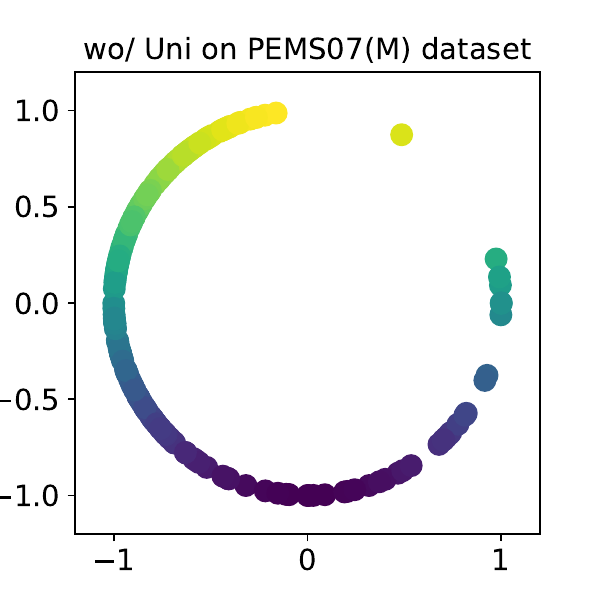}
    \end{minipage}%
}%

\centering
\vspace{-0.35cm}
\caption{Distribution of the prompt embedding.}
\vspace{-0.35cm}
\label{fig:case}
\end{figure}
\section{Conclusion}
\label{sec:conclusion}
\vspace{-0.05in}

This paper presents \model, a new and efficient method for adapting spatio-temporal predictive models to various downstream tasks that involve previously unseen data. Our in-context learning framework utilizes a spatio-temporal prompt network, which consists of a context distillation mechanism and a dependency modeling scheme. This framework effectively adapts to different spatio-temporal scenarios by capturing contextual signals and modeling complex relationships across time and locations. To address the distribution gap, we enhance \model\ by incorporating a distribution mapping mechanism that aligns the data distributions of pre-training and downstream data, facilitating effective knowledge transfer in spatio-temporal forecasting. Extensive experiments demonstrate the effectiveness and generalization capabilities of our \model\ in diverse downstream spatio-temporal forecasting scenarios. For future work, an interesting direction would be to explore the potential of incorporating large language models as knowledge guidance within our \model\ framework.

\section{Impact Statements}
\label{sec:Impact}
This paper presents work whose goal is to advance the field of Machine Learning. There are many potential societal consequences of our work, none which we feel must be specifically highlighted here.

\section*{Acknowledgments}
\label{sec:Acknowledgments}
This work is partially supported by the Guangzhou Municipal Science and Technology Plan Project - Key Research and Development Program (Grant No. 2024B01W0007).


\clearpage


\bibliography{refs}
\bibliographystyle{icml2024}






\end{document}